\documentclass[10pt,twocolumn,letterpaper]{article}

\usepackage[pagenumbers]{cvpr} 
\usepackage[accsupp]{axessibility} 

\usepackage{times}
\usepackage{epsfig}
\usepackage{graphicx}
\usepackage{amsmath}
\usepackage{amssymb}
\usepackage{booktabs}
\usepackage{tabu}
\usepackage{soul}
\usepackage{enumitem}
\usepackage{bbm}
\usepackage{mathtools}
\usepackage{xurl}
\usepackage{diagbox}
\usepackage{multirow}
\usepackage[normalem]{ulem}
\usepackage[table,xcdraw]{xcolor}
\usepackage{stfloats}
\usepackage{pifont}

\useunder{\uline}{\ul}{}

\usepackage[pagebackref,breaklinks,colorlinks]{hyperref}

\usepackage[capitalize]{cleveref}
\crefname{section}{Sec.}{Secs.}
\Crefname{section}{Section}{Sections}
\Crefname{table}{Table}{Tables}
\crefname{table}{Tab.}{Tabs.}

\def\cvprPaperID{15} 
\def\confName{CVPR}
\def\confYear{2023}

\begin{document}

\title{Robust Multiview Multimodal Driver Monitoring System Using Masked Multi-Head Self-Attention}

\author{Yiming Ma\\
University of Warwick\\
Coventry, UK\\
\and
Victor Sanchez\\
University of Warwick\\
Coventry, UK\\
\and
Soodeh Nikan\\
Ford Motor Company\\
USA\\
\and
Devesh Upadhyay\\
Ford Motor Company\\
USA\\
\and
Bhushan Atote\\
University of Warwick\\
Coventry, UK\\
\and
Tanaya Guha\\
University of Glasgow\\
Glasgow, UK\\
}
\maketitle
\thispagestyle{empty}

\begin{abstract}
Driver Monitoring Systems (DMSs) are crucial for safe hand-over actions in Level-2+ self-driving vehicles. State-of-the-art DMSs leverage multiple sensors mounted at different locations to monitor the driver and the vehicle's interior scene and employ decision-level fusion to integrate these heterogenous data. However, this fusion method may not fully utilize the complementarity of different data sources and may overlook their relative importance. To address these limitations, we propose a novel multiview multimodal driver monitoring system based on feature-level fusion through multi-head self-attention (\textbf{MHSA}). We demonstrate its effectiveness by comparing it against four alternative fusion strategies (Sum, Conv, SE, and AFF). We also present a novel GPU-friendly supervised contrastive learning framework \textbf{SuMoCo} to learn better representations. Furthermore, We fine-grained the test split of the DAD dataset to enable the multi-class recognition of drivers' activities. Experiments on this enhanced database demonstrate that 1) the proposed MHSA-based fusion method (AUC-ROC: 97.0\%) outperforms all baselines and previous approaches, and 2) training MHSA with patch masking can improve its robustness against modality/view collapses. The code and annotations are publicly available \footnote{\url{https://github.com/Yiming-M/MHSA}}.
\end{abstract}

\section{Introduction}
\label{sec:1}
{
\definecolor{myorange}{HTML}{ED723F}
\definecolor{mygreen}{HTML}{215A59}
\begin{figure}[t]
    \centering
    \includegraphics[width=\linewidth]{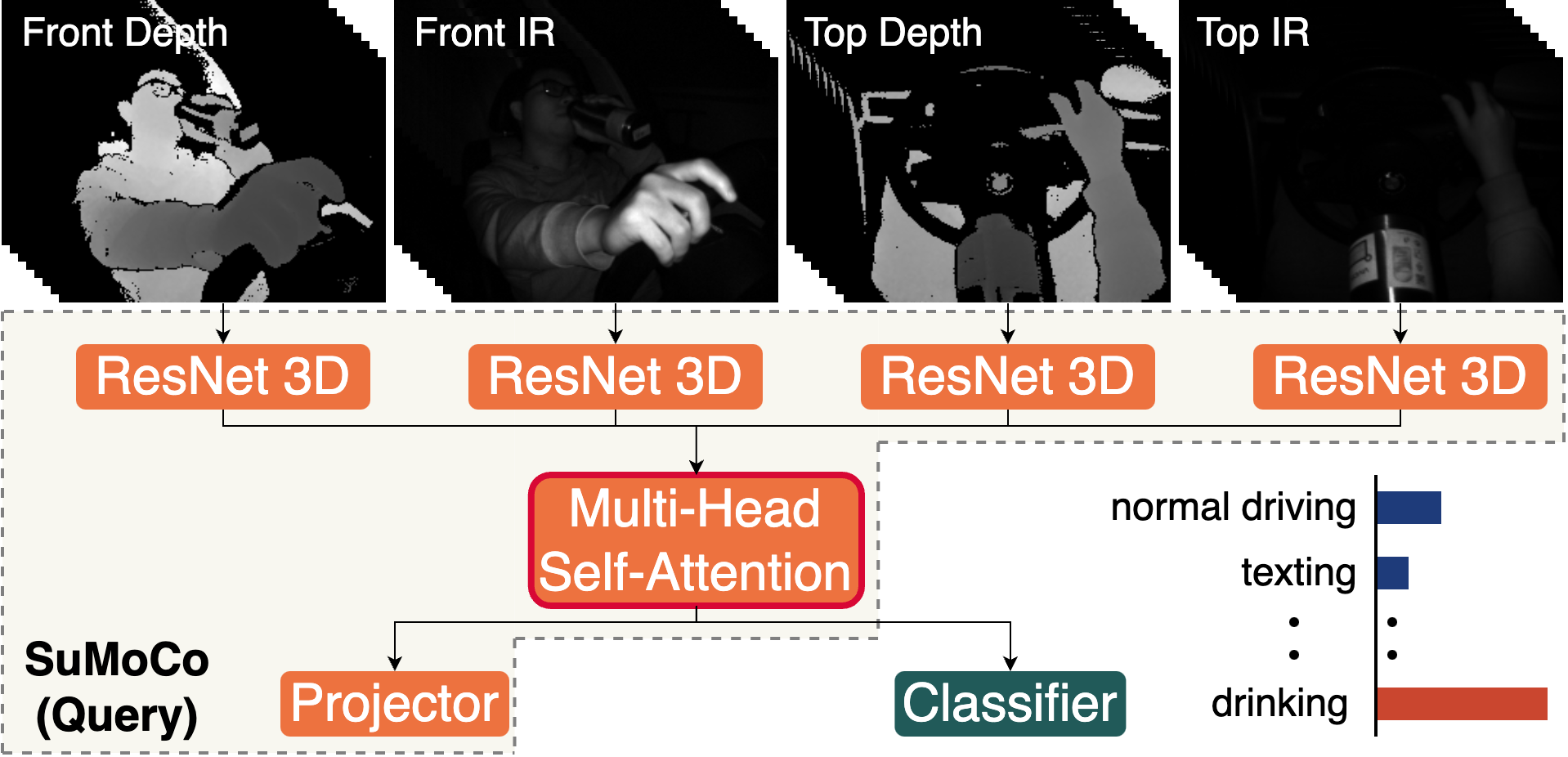}
    \caption{An overview of our proposed DMS: R3D-18 \cite{tran2018closer} backbones are utilized to extract spatial-temporal features from the multiview multimodal inputs. These feature maps are subsequently fused via multi-head self-attention ( illustrated in Figure~\ref{fig:mhsa}). A 2-layer perceptron is leveraged to project the fused features into the contrastive embedding sapce, while another 2-layer perceptron generates the score for each class. The \textcolor{myorange}{orange} blocks constitute the query encoder of our proposed contrastive learning framework, SuMoCo. They are supervised using the infoNCE loss \cite{oord2018representation}, and the \textcolor{mygreen}{classifier} is trained separately using the focal loss \cite{lin2017focal}.}
    \label{fig:base_structure}
\end{figure}
}
Modern \textit{driver monitoring systems} (DMSs) in Level-2+ self-driving-enabled cars aim to enhance safety by estimating drivers' readiness levels for driving and enabling safe control handovers when necessary. These systems usually rely on various visual sensors installed at different locations within the car to monitor drivers' states comprehensively. For instance, cameras installed above the driver can collect data related to hand-involved activities (e.g., messaging). Those in front can monitor the driver's upper body movements, enabling the detection of actions like drinking. While the RGB modality provides sufficient optical details for object detection, near infrared (NIR) can enhance robustness under adverse environmental conditions such as poor lighting. Given these \textit{multiview} \textit{multimodal} data, effectively integrating them is thus crucial for DMSs to become applicable in the real world.

The study of DMSs encompasses various domains, and this paper focuses specifically on \textit{driver action recognition}. This task involves classifying a driver's actions into ``\textit{normal driving}'' and several \textit{non-driving-related activities} (NDRAs) such as talking and drinking. Existing approaches \cite{kopuklu2021driver, ortega2020dmd} typically employ a naive fusion method that combines the multiview multimodal data at the decision level. However, this approach fails to exploit the complementarity of the semantic features from different views and modalities and does not consider their relative importance, leading to underperformance. Hence, we propose DMSs based on feature-level fusion with self-attention, as depicted in Fig.~\ref{fig:base_structure}, to address these limitations. Following the prior work \cite{kopuklu2021driver}, we also train our models with supervised \textit{contrastive learning} (CL). To this end, we introduce \textbf{SuMoCo}, a novel GPU-friendly framework based on its unsupervised counterpart, MoCo \cite{he2020momentum}.

Given that drivers can perform indefinite non-driving-related activities, we evaluate our proposed methods on the Driver Anomaly Detection Dataset (DAD) \cite{kopuklu2021driver}, which is designed explicitly for open-set NDRA detection. The dataset consists of only two categories, namely ``normal'' and ``anomalous'' (the class for all NDRAs). However, identifying the specific type of NDRA is critical in practice as they pose varying degrees of risk from inattention. Therefore, we manually annotate DAD with fine-grained labels, such as ``drinking'', ``talking'' and "texting", to enable the multi-classification of NDRAs.


{
\begin{figure*}[t]
    \centering
     \includegraphics[width=\textwidth]{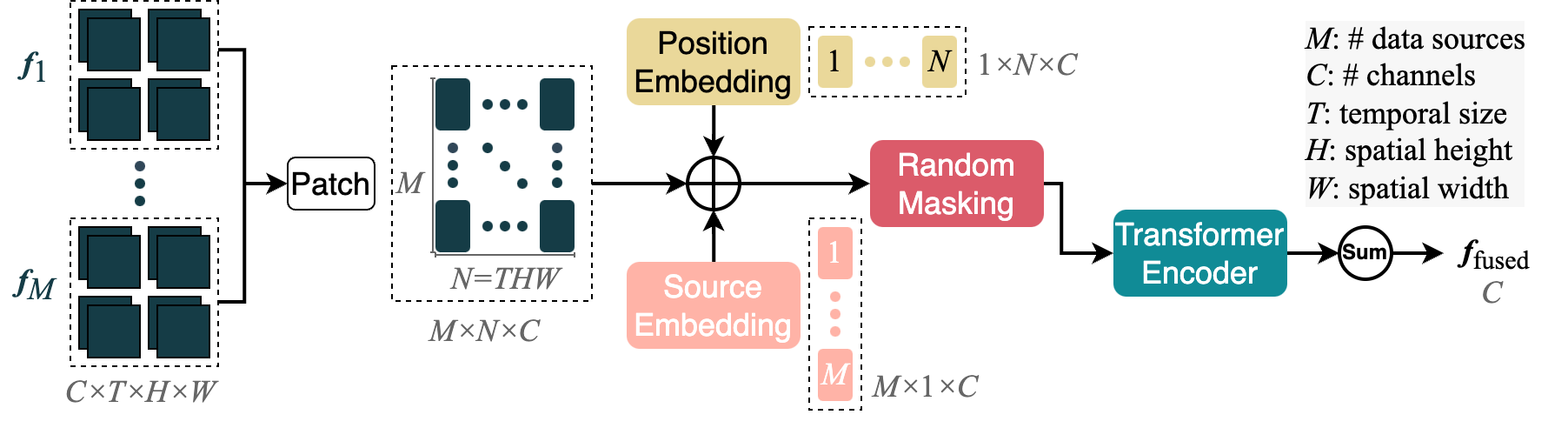}
     \caption{The structure of our proposed multi-head self-attention feature fusion module \textbf{MHSA}. We first split the extracted feature maps into fixed-size patches and add the source embedding and the positional embedding to them. Next, we randomly mask some patches and feed the remaining into the transformer encoder block \cite{vaswani2017attention, dosovitskiy2021an} to learn interactions among them and capture the global context. This masking operation simulates view/modality collpase, leading to improved robustness. Finally, the attended patches are summed to generate the output.}
     \label{fig:mhsa}
\end{figure*}
}

Our contributions in this paper are three-fold as follows:
\begin{enumerate}[itemsep=2pt,topsep=0pt,parsep=0pt]
    \item We present a novel multiview multimodal driver monitoring system (DMS) that leverages feature-level fusion through multi-head self-attention (\textbf{MHSA}). To demonstrate the effectiveness of our proposed fusion method, we introduce four alternative fusion strategies: \textbf{Sum}, \textbf{Conv}, Squeeze-and-Excitation (\textbf{SE}), and Attentional Feature Fusion (\textbf{AFF}). We propose a new supervised contrastive learning framework, \textbf{SuMoCo}, to efficiently train these models.
    \item We manually annotated the DAD dataset with the specific labels of non-driving-related activities (NDRAs) to enable their recognition. Consequently, in the test set, the ``anomalous'' class is replaced by nine fine-grained classes. These newly introduced labels offer greater granularity and thus have the potential to enhance the identification of the most distracting NDRAs. These additional annotations have been made publicly available.
    \item We conduct extensive experiments on the DAD dataset to compare different fusion strategies, assess the significance of individual views/modalities, and evaluate the efficacy of patch masking in enhancing MHSA's robustness against view/modality collapses. Results show that our MHSA-based DMS achieves state-of-the-art performance with an AUC-ROC score of 97.0\%. 
\end{enumerate}

\section{Related Work}
\label{sec:2}

\subsection{Multimodal Driver Monitoring Systems}
\noindent \textbf{Datasets:} StateFarm \cite{statefarm} and AUC-DD \cite{abouelnaga2018real} were among the earliest datasets for driver action recognition. They were collected using an RGB camera from a single side view and thus have some limitations. For instance, certain hand-related activities (e.g., texting) may be occluded, and the RGB camera cannot provide sufficient optical details in poor illumination conditions. Thus, methods \cite{abouelnaga2018real, baheti2018detection, eraqi2019driver} developed on these datasets may not be robust enough for practical use.

Later databases \cite{martin2019drive, ortega2020dmd, jegham2020novel, kopuklu2021driver} have incorporated additional views and modalities to address these issues. For example, top and front views have also been introduced to capture the driver's hand and head movements amongst other movements. Regarding modalities, IR and depth have also become popular, as they can provide thermally based features and geometry information, which are complementary to the optical details from RGB. Among these datasets, we benchmark our models on DAD \cite{kopuklu2021driver}, the only one designed for SAE L2+ with open-set recognition: its test set contains extra classes of NDRAs in addition to those in the training split. This characteristic makes it representative of the real-world driving scenario, where there can be unboundedly many types of NDRAs.

\noindent\textbf{Multimodal DMSs.} Various multiview multimodal methods have been proposed with different emphases. Some propose novel learning methods (e.g., supervised contrastive learning \cite{kopuklu2021driver}), while others \cite{ortega2020dmd, canas2021detection, su2022efficient, shan2022multi, abdullah2022multi} are focused on handling the temporal dimension. However, how to combine heterogenous data in DMS has rarely been studied. Most previous methods \cite{kopuklu2021driver, martin2019drive, su2022efficient} adopt a decision-level fusion by averaging the scores, while Ortega \textit{et al.} \cite{ortega2020dmd} propose to fuse data at an input level by concatenation. These strategies cannot handle modality/view interaction well and hence tend to underperform. The former neglects the extracted feature maps that can correctly describe the driver's actions only when compared and combined, while latter ignores the spatial inconsistency when concatenating all input videos along the channel dimension. Only Shan \textit{et al.} \cite{shan2022multi} propose a nontrivial multimodal approach, but it has several drawbacks: 1) features are pooled before fusion, which leads to the loss of semantic information; and 2) its fusion module has the additional task of handling the temporal dimension. By contrast, our work is the first to specifically investigate how to effectively fuse modalities and views at the feature level in driver monitoring systems.

\subsection{Contrastive Learning}

Kopuklu \textit{et al}. \cite{kopuklu2021driver} propose a supervised contrastive learning method for NDRA detection on DAD. This method and the state-of-the-art supervised contrastive learning method SupContrast \cite{khosla2020supervised} are both based on the unsupervised framework SimCLR \cite{chen2020simple}, which requires large batch sizes to estimate the infoNCE loss \cite{oord2018representation} accurately. For instance, in \cite{kopuklu2021driver}, each batch comprises 160 clips with a size each of $16 \times 112 \times 112$. This large input size makes these CL methods impractical, as they require huge GPU memory to calculate gradients. By comparison, MoCo \cite{he2020momentum} contrast the current extracted embeddings with those previous ones stored in a queue. to address this issue. Besides, MoCo optimizes the key encoder's weights with a momentum-based update scheme to guarantee the consistency of embeddings extracted by it. However, it is based on unsupervised learning and needs to be adapted for the supervised scenario. We fill this gap by caching both embeddings and labels into the queue. Specifically, our framework groups embeddings with the same labels together and separates those with different labels.

\begin{figure}[t]
    \centering
    \includegraphics[width=\linewidth]{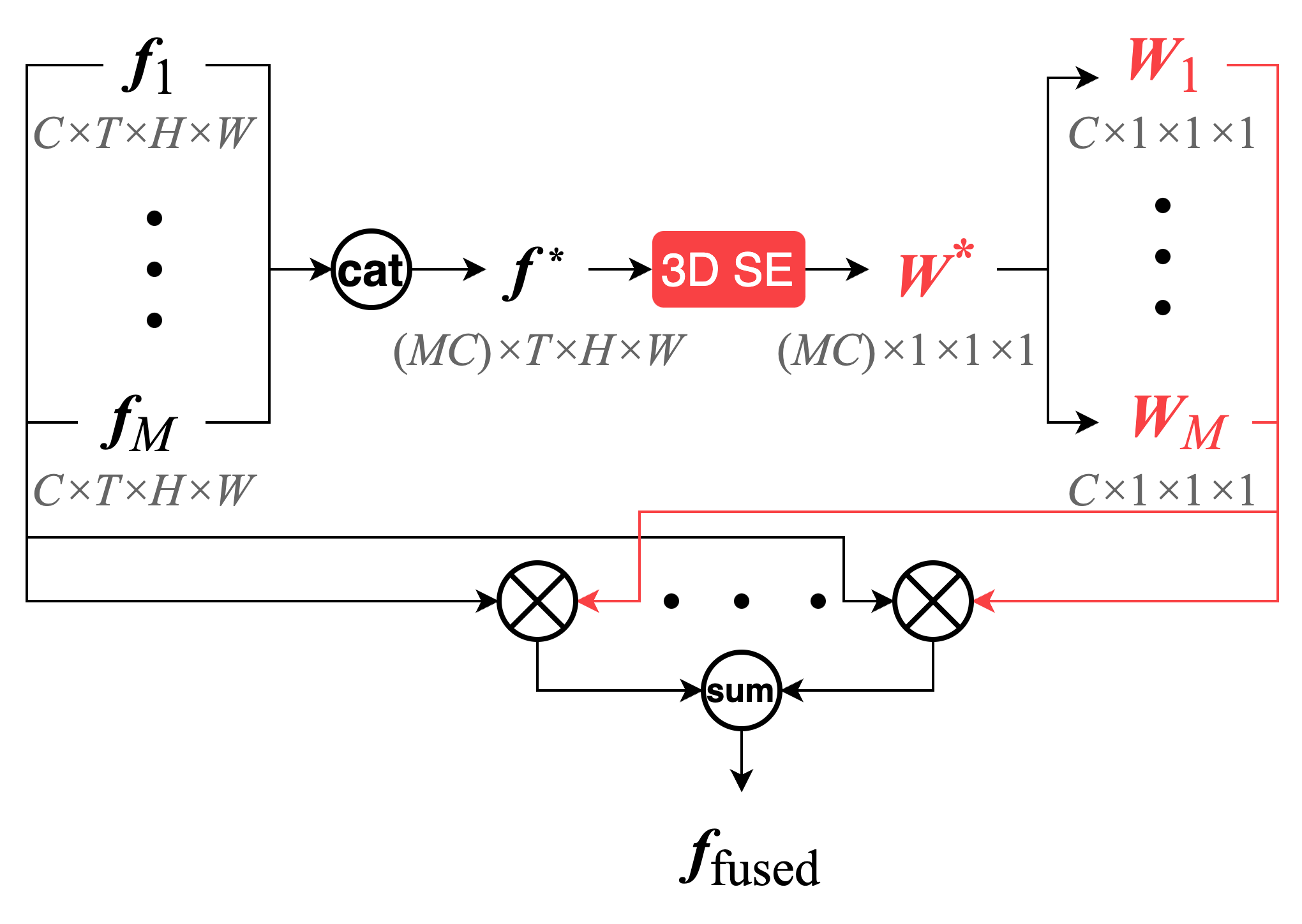}
    \caption{The structure of our proposed squeeze-and-excitation feature fusion module \textbf{SE}. Feature maps are first concatenated along the channel axis. We leverage the squeeze-and-excitation mechanism \cite{hu2018squeeze} to learn the weight for each channel. The weight matrices are then used to average the input feature maps. Through this way, our method can model the interaction between different views and modalities and learn the corresponding relative importance.}
    \label{fig:other_fusion_method}
\end{figure}

\section{Method}
\label{sec:3}

We propose a multiview multimodal DMS that employs feature-level fusion (see Fig. \ref{fig:base_structure}). Let $\{ \boldsymbol{X}_1, \, \cdots \, \boldsymbol{X}_M \}$ be the input video clips collected at the same time from $M$ different sources. Since data from different sources have distinct statistical distributions, for the video clip $\boldsymbol{X}_i$, we use a separate R3D-18 \cite{tran2018closer} backbone $\boldsymbol{F}_i$ to extract the feature map $\boldsymbol{f}_i$. Specifically, for $i = 1, \, \cdots, \, M$, we have
\begin{equation} \label{eqn:1}
    \boldsymbol{f}_{i} = \boldsymbol{F}_{i} \left( \boldsymbol{X}_{i} \right) \in \mathbb{R}^{C \times T \times H \times W},
\end{equation}
where $C$ is the number of channels in $\boldsymbol{f}$, $T$ denotes the temporal dimension size,  and $H$ and $W$ refer to the height and width of the spatial dimension.


\begin{table*}[t]
\definecolor{shade}{HTML}{FFFFC7}
\definecolor{best}{HTML}{ff0054}
\sethlcolor{shade}
\resizebox{\textwidth}{!}{
\begin{tabular}{l|cccccc|cccccc|cccccc}
\toprule
 &
  \multicolumn{6}{c|}{\textbf{Top}} &
  \multicolumn{6}{c|}{\textbf{Front}} &
  \multicolumn{6}{c}{\textbf{Top + Front}} \\ \cline{2-19} 
 &
  \multicolumn{2}{c|}{\textbf{D}} &
  \multicolumn{2}{c|}{\textbf{IR}} &
  \multicolumn{2}{c|}{\textbf{D + IR}} &
  \multicolumn{2}{c|}{\textbf{D}} &
  \multicolumn{2}{c|}{\textbf{IR}} &
  \multicolumn{2}{c|}{\textbf{D + IR}} &
  \multicolumn{2}{c|}{\textbf{D}} &
  \multicolumn{2}{c|}{\textbf{IR}} &
  \multicolumn{2}{c}{\textbf{D + IR}} \\ \cline{2-19} 
\multirow{-3}{*}{\textbf{Fusion}} &
  \textbf{ROC} &
  \multicolumn{1}{c|}{\textbf{mAP}} &
  \textbf{ROC} &
  \multicolumn{1}{c|}{\textbf{mAP}} &
  \textbf{ROC} &
  \textbf{mAP} &
  \textbf{ROC} &
  \multicolumn{1}{c|}{\textbf{mAP}} &
  \textbf{ROC} &
  \multicolumn{1}{c|}{\textbf{mAP}} &
  \textbf{ROC} &
  \textbf{mAP} &
  \textbf{ROC} &
  \multicolumn{1}{c|}{\textbf{mAP}} &
  \textbf{ROC} &
  \multicolumn{1}{c|}{\textbf{mAP}} &
  \textbf{ROC} &
  \textbf{mAP} \\ \midrule
  Decision \cite{kopuklu2021driver} &
  91.3 &
  \multicolumn{1}{c|}{-} &
  88.0 &
  \multicolumn{1}{c|}{-} &
  {91.7} &
  - &
  90.0 &
  \multicolumn{1}{c|}{-} &
  87.0 &
  \multicolumn{1}{c|}{-} &
  \cellcolor{shade}{92.0} &
  - &
  \cellcolor{shade}{96.1} &
  \multicolumn{1}{c|}{-} &
  \cellcolor{shade}93.1 &
  \multicolumn{1}{c|}{-} &
  \cellcolor{shade}96.6 &
  - \\ \midrule
Sum &
   &
  \multicolumn{1}{c|}{} &
   &
  \multicolumn{1}{c|}{} &
  91.7 &
  94.2 &
   &
  \multicolumn{1}{c|}{} &
   &
  \multicolumn{1}{c|}{} &
  \cellcolor{shade}{92.7} &
  93.2 &
  \cellcolor{shade}{94.8} &
  \multicolumn{1}{c|}{\cellcolor{shade}{96.8}} &
  \cellcolor{shade}94.5 &
  \multicolumn{1}{c|}{\cellcolor{shade}96.0} &
  \cellcolor{shade}96.3 &
  \cellcolor{shade}96.8 \\
Conv &
   &
  \multicolumn{1}{c|}{} &
   &
  \multicolumn{1}{c|}{} &
  92.2 &
  94.3 &
   &
  \multicolumn{1}{c|}{} &
   &
  \multicolumn{1}{c|}{} &
  \cellcolor{shade}{92.9} &
  94.1 &
  \cellcolor{shade}{95.8} &
  \multicolumn{1}{c|}{\cellcolor{shade}{97.4}} &
  \cellcolor{shade}94.6 &
  \multicolumn{1}{c|}{\cellcolor{shade}96.1} &
  \cellcolor{shade}96.2 &
  \cellcolor{shade}97.5 \\
SE &
   &
  \multicolumn{1}{c|}{} &
   &
  \multicolumn{1}{c|}{} &
  92.3 &
  94.3 &
   &
  \multicolumn{1}{c|}{} &
   &
  \multicolumn{1}{c|}{} &
  \cellcolor{shade}{92.9} &
  \cellcolor{shade}{94.5} &
  \cellcolor{shade}{95.9} &
  \multicolumn{1}{c|}{\cellcolor{shade}{97.4}} &
  \cellcolor{shade}94.9 &
  \multicolumn{1}{c|}{\cellcolor{shade}96.5} &
  \cellcolor{shade}96.4 &
  \cellcolor{shade}97.6 \\
AFF &
   &
  \multicolumn{1}{c|}{} &
   &
  \multicolumn{1}{c|}{} &
  92.5 &
  94.7 &
   &
  \multicolumn{1}{c|}{} &
   &
  \multicolumn{1}{c|}{} &
  \cellcolor{shade}{\textbf{93.1}} &
  \cellcolor{shade}{94.7} &
  \cellcolor{shade}{96.4} &
  \multicolumn{1}{c|}{\cellcolor{shade}{97.6}} &
  \cellcolor{shade}95.0 &
  \multicolumn{1}{c|}{\cellcolor{shade}96.6} &
  \cellcolor{shade}96.7 &
  97.4 \\
  \textbf{MHSA (ours)}&
  \multirow{-5}{*}{\textbf{92.9}} &
  \multicolumn{1}{c|}{\multirow{-5}{*}{94.9}} &
  \multirow{-5}{*}{\textbf{91.3}} &
  \multicolumn{1}{c|}{\multirow{-5}{*}{93.5}} &
  \cellcolor{shade}{\textbf{92.9}} &
  \cellcolor{shade}{\textbf{95.2}} &
  \multirow{-5}{*}{\textbf{91.7}} &
  \multicolumn{1}{c|}{\multirow{-5}{*}{94.4}} &
  \multirow{-5}{*}{\textbf{90.2}} &
  \multicolumn{1}{c|}{\multirow{-5}{*}{91.8}} &
  \cellcolor{shade}{\textbf{93.1}} &
  \cellcolor{shade}{\textbf{94.9}} &
  \cellcolor{shade}{\textbf{96.7}} &
  \multicolumn{1}{c|}{\cellcolor{shade}{\textbf{97.7}}} &
  \cellcolor{shade}\textbf{95.7} &
  \multicolumn{1}{c|}{\cellcolor{shade}\textbf{97.1}} &
  \cellcolor{shade}{\color{best}\textbf{97.0}} &
  \cellcolor{shade}{\color{best}\textbf{97.8}} \\ \bottomrule
\end{tabular}}
\caption{Results for NDRA detection on the DAD dataset. Each samples is classified as either ``normal driving'' or not. Here, \textbf{D} represents the depth modality and \textbf{IR} the infrared modality. The highest score for each view and modality are in \textbf{bold}, and those \hl{highlighted} indicate improvement brought by introducing an extra modality or view.}
\label{tbl:binary}
\end{table*}

\subsection{Multi-Head Self-Attention Fusion}

We propose a novel fusion method \textbf{MHSA} (see Fig.~\ref{fig:mhsa}), which is based on the multi-head self-attention. First, we divide each feature map $\boldsymbol{f}_{i} \in \mathbb{R}^{C \times T \times H \times W}$ obtained by Eq. \eqref{eqn:1} into patches of size $C \times 1 \times 1 \times 1$, resulting in $N := T H W$ patches per source, denoted by 
$$
\left\{\boldsymbol{p}^{(j)}_i \in \mathbb{R}^{C} \, \middle| \, j = 1, \, \cdots, \, N, \, i = 1, \, \cdots , \, M \right\}.
$$
Then, we infuse the source embedding $\boldsymbol{m}_{i} \in \mathbb{R}^{C}$ and the positional embedding $\boldsymbol{s}^{(j)} \in \mathbb{R}^{C}$ into each patch $\boldsymbol{p}^{(j)}_i$ via addition:
$$
\boldsymbol{q}^{(j)}_{i} = \boldsymbol{p}^{(j)}_{i} + \boldsymbol{m}_{i} + \boldsymbol{s}^{(j)}.
$$
The resulting patches $\boldsymbol{q}^{(j)}_{i}$ for $j = 1, \, \cdots, \, N$ and $i = 1, \, \cdots , \, M$ thus preserving information about their data sources and the spatiotemporal positions in the original feature maps.

During training, we randomly mask $n\%$ of the patches $\{ \boldsymbol{q}^{(j)}_{i} | j = 1, \, \cdots, \, N, \, i = 1, \, \cdots , \, M \}$ to enhance the model's robustness against corrupt modalities or views. After this step, the multi-head self-attention mechanism \cite{vaswani2017attention, dosovitskiy2021an} is applied to the remaining patches $\{ \boldsymbol{q}_1, \, \cdots, \, \boldsymbol{q}_r \}$, where $r$ is the number of unmasked patches. This mechanism distributes attention to patches from different sources and spatiotemporal positions and can thus learn the relative importance of each patch. Finally, the output patches $\{ \boldsymbol{u}_1, \, \cdots, \, \boldsymbol{u}_r \}$ are combined via addition to obtain a global representation:
\begin{equation} \label{eqn:2}
    \boldsymbol{g} = \sum_{i=1}^r \boldsymbol{u}_i \in \mathbb{R}^{C}.
\end{equation}

It is worth noting that MHSA is different from the fusion mechanism  of Shan \textit{et al.} \cite{shan2022multi} in the following aspects.
\begin{enumerate}[itemsep=2pt, topsep=0pt, parsep=0pt]
    \item MHSA \textit{focuses on view/modality interaction}. We use 3D CNN backbones to extract spatial-temporal features. In contrast, \cite{shan2022multi} leverages 2D CNN, so their fusion mechanism needs also to handle the temporal dimension: attention is distributed across modalities and temporal steps.
    \item MHSA \textit{preserves input semantics}. Features are NOT pooled to generate patches, and also, the source encoding and the positional encoding are introduced to preserve the data sources and spatiotemporal positions of patches.
    \item MHSA is \textit{more efficient}. To obtain a good global representation, MHSA simply adds the attended features, while Shan \textit{et al.} uses a class token and multiple chained transformer blocks.
\end{enumerate}

\subsection{Other Fusion Methods}

To compare with our proposed MHSA fusion module, we propose four alternative feature fusion methods,since similar approaches have not been explored on DMSs before.

\vspace{1em}
\noindent\textbf{Sum.} This most straightforward method fuses the four feature maps by directly adding them:
\begin{equation} \label{eqn:3}
    \boldsymbol{f} = \sum_{i=1}^M \boldsymbol{f}_{i}.
\end{equation}

\noindent\textbf{Conv.} Features are first concatenated along the channel dimension, as follows:
\begin{equation} \label{eqn:4}
    \boldsymbol{f}^* = \left[ \boldsymbol{f}_i \, \| \cdots \| \boldsymbol{f}_M \right] \in \mathbb{R}^{MC \times T \times H \times W}.
\end{equation}
Then a point-wise convolution is performed to reduce the channel size to $C$:
\begin{equation} \label{eqn:5}
    \boldsymbol{f} = \text{Conv} \left( \boldsymbol{f}^* \right).
\end{equation}

\noindent\textbf{SE.} Figure~\ref{fig:other_fusion_method} depicts the structure of this module. The 3D version of squeeze-and-excitation \cite{hu2018squeeze} is imposed on the concatenated feature maps in Eq. \eqref{eqn:4} to learn the channel attention matrix:
\begin{equation} \label{eqn:6}
    \boldsymbol{W}^* = \text{SE} \left( \boldsymbol{f^*} \right) \in \mathbb{R}^{MC \times 1 \times 1 \times 1},
\end{equation}
which is then split along the channel axis into $M$ chunks $\boldsymbol{W}_{i} \in \mathbb{R}^{C \times 1 \times 1 \times 1}$, for $i = 1, \, \cdots , \, M$. Finally, features are fused by weighted average:
\begin{equation} \label{eqn:7}
    \boldsymbol{f} = \sum_{i=1}^M \boldsymbol{f}_{i} * \boldsymbol{W}_{i},
\end{equation}
where ``$*$'' represents the element-wise prodcut.\\

\vspace{-2mm}
\noindent \textbf{AFF.} This module is similar to \textbf{SE}, but instead of using squeeze-and-excitation, it utilizes the 3D Attentional Feature Fusion module \cite{dai2021attentional} to learn both spatial-temporal attention and channel-wise attention. Specifically, the attention matrix is calculated as follows:
\begin{equation} \label{eqn:8}
    \boldsymbol{W}^* = \text{AFF} \left( \boldsymbol{f}^* \right) \in \mathbb{R}^{MC \times T \times H \times W},
\end{equation}
which is then chunked to average $\boldsymbol{f}_{i}$ like in Eq. \eqref{eqn:7}.

After fusing features by \eqref{eqn:3}, \eqref{eqn:5} or \eqref{eqn:7}, an average pooling layer and a flatten layer are utilized to transform $\boldsymbol{f} \in \mathbb{R}^{C \times T \times H \times W}$ into a vector:
\begin{equation} \label{eqn:9}
    \boldsymbol{g} = \text{Flatten} \left( \text{Avg}  \left( \boldsymbol{f} \right) \right) \in \mathbb{R}^{\times C}.
\end{equation}

\begin{table}[t]
\centering
\resizebox{\columnwidth}{!}{%
\begin{tabular}{l|cccc}
\hline
\textbf{CL Framework}    & \textbf{Top D} & \textbf{Top IR} & \textbf{Front D} & \textbf{Front IR} \\ \hline
Kopuklu \textit{et al}. \cite{kopuklu2021driver} & 91.3 & 88.0 & 90.0 & 87.0 \\ \hline
SuMoCo (w/o new labels)                         & 90.8 & 89.8 & 89.9 & 88.7 \\
SuMoCo (w/ new labels) & \textbf{92.9}  & \textbf{91.3}   & \textbf{91.7}    & \textbf{90.2}     \\ \hline
\end{tabular}%
}
\caption{Results comparing our contrastive learning framework SuMoCo with Kopuklu \textit{et al}. \cite{kopuklu2021driver} in single-view single-modal NDRA detection, as measured by the AUC-ROC metric. Initially, SuMoCo can only perform binary classification without our labels. However, after manually annotating the test set, it can be trained for multi-classification, and its results for detection are shown in the last row.}
\label{tbl:sumoco}
\end{table}

\subsection{SuMoCo: A Novel Supervised Contrastive Learning Framework}

We introduce a novel supervised momentum contrastive learning framework \textbf{SuMoCo}, based on MoCo \cite{he2020momentum} and SupContrast \cite{khosla2020supervised}. Like its self-supervised counterpart, SuMoCo also comprises a query encoder $\mathcal{E}_Q$ and a query projection head $\mathcal{P}_Q$, which are then copied to initialize the key encoder $\mathcal{E}_K$ and the key projection head $\mathcal{P}_K$. The encoder $\mathcal{E}$ is composed of R3D-18 backbones and fusion modules, and produces an output $\boldsymbol{g} \in \mathbb{R}^{C}$ (determined by either \eqref{eqn:2} or \eqref{eqn:9}). We use a two-layer perceptron as the projection head $\mathcal{P}$.

For each mini batch, we first compute the contrastive embeddings $\boldsymbol{z} = \mathcal{P} (\mathcal{E} (\boldsymbol{X}))$. Subsequently, the embeddings from the key encoder $\boldsymbol{z}_K$ are detached from the gradient graph and stored in the queue with their corresponding labels $\boldsymbol{y}$. The weights of the query encoder $\mathcal{E}_Q$ and projection head $\mathcal{P}_Q$ are updated by the infoNCE loss \cite{oord2018representation}, defined by the following equation:
\begin{equation} \label{eqn:10}
    \mathcal{L} = - \sum_{i=1}^B \sum_{p \in P(i)} \log \frac{\exp{(\boldsymbol{z}^{(i)}_Q \cdot \boldsymbol{z}^{(p)}_K / \tau)}}{\sum_{a \in A(i)} \exp{(\boldsymbol{z}^{(i)}_Q \cdot \boldsymbol{z}^{(a)}_K / \tau)}},
\end{equation}
where $B$ is the batch size, $P(i)$ is the set of instances in the queue that has the same label as $i$'s, $A(i)$ is the set of instances with opposite labels, and $\tau$ is a temperature parameter controlling the distribution of the embeddings. To ensure consistent output, the weights of $\mathcal{E}_K$ and $\mathcal{P}_K$ are updated with a momentum $m$ :
\begin{equation} \label{eqn:11}
    \boldsymbol{\mathcal{W}}_K = m \cdot \boldsymbol{\mathcal{W}}_K + (1-m) \cdot \boldsymbol{\mathcal{W}}_Q.
\end{equation}

Next, we detach the query embeddings $\boldsymbol{g}_Q$ from the gradient graph and feed them to a two-layer classifier to generate scores for each class. The focal loss \cite{lin2017focal} supervises the training of this prediction head.

\begin{figure}[t]
    \centering
    \includegraphics[width=\linewidth]{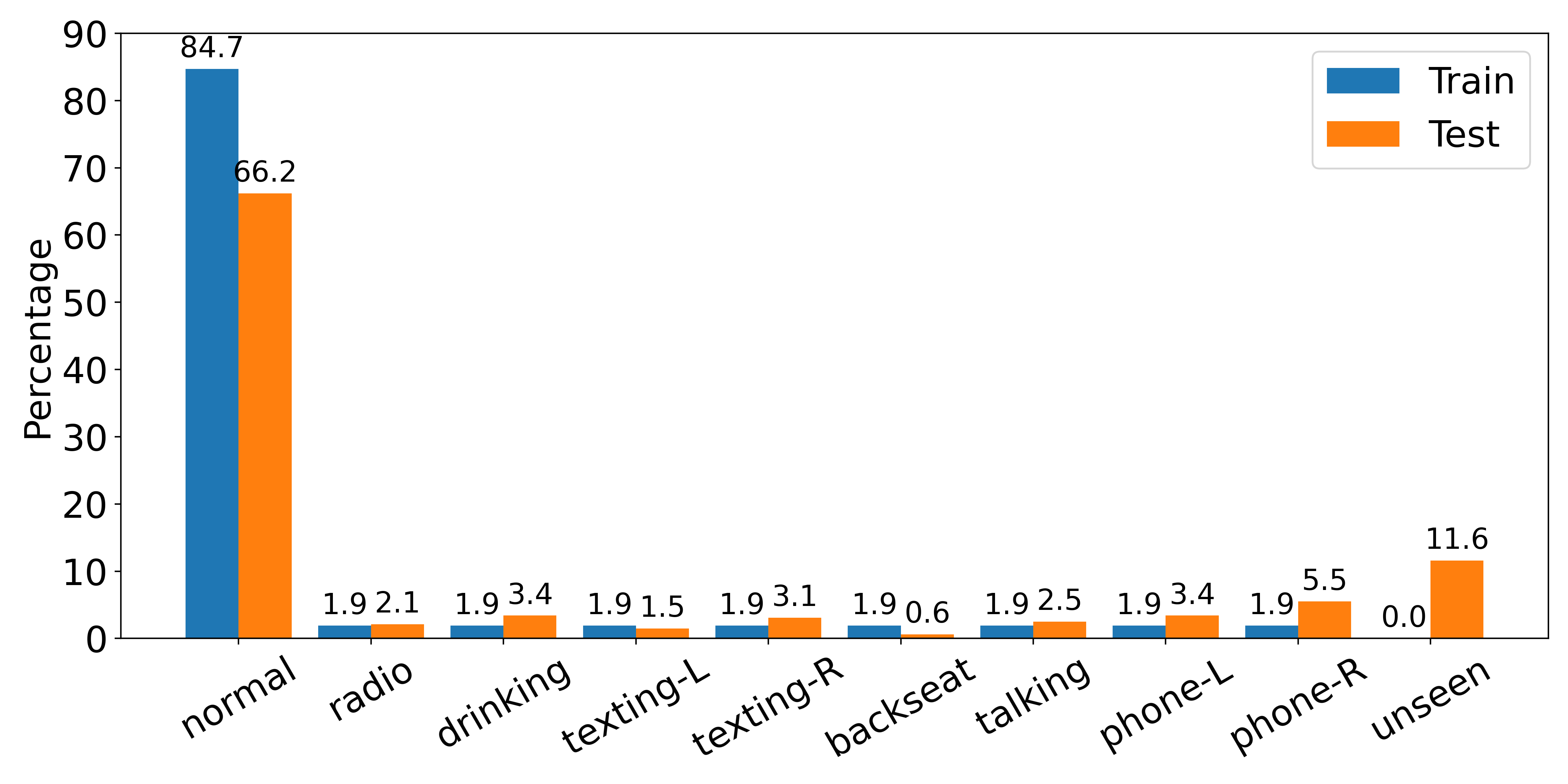}
    \caption{The distribution of the fine-grained classses. The label ``normal'' refers to normal driving, and the other nine are non-driving-ralated activities: ``radio'': tuning the radio; ``backseat'': reaching the back seat; ``talking'': talking with the passenger; ``phone'': talking on the phone. The ``L/R'' suffix stand for with the left/right hand, and those NDRAs only in test set are labeled as ``unseen''.
    }
    \label{fig:class_distribution}
\end{figure}

\section{Experiments}
\label{sec:4}

\noindent\textbf{New Annotations.} We evaluate our work based on the DAD databse \cite{kopuklu2021driver}, which was designed for NDRA detection. Since the set of all possible actions performed by drivers is unbounded, its test split contains more types of NDRA than the training set to better estimate real-world performances of DMSs. However, all the NDRAs in the test set are labeled as ``anomlaous driving'' instead of their specific types, hindering DMSs from classifying them. On the other hand, recognizing the specific activities is of crucial importance in practice, as different unrelated activities require varying amounts of response time for drivers to refocus their attention on driving. To bridge this gap, we have manually annotated each sample in the test set with its corresponding label.

\noindent\textbf{Dataset Statistics.} The DAD database \cite{kopuklu2021driver} was collected at 47 FPS from two views (top \& front) and two modalities (IR \& depth). Its training set comprises 1,770,000 frames for each data source, and its test set contains 360,000 frames. The training set has one class for normal driving and eight NDRA categories, while the test split has one additional class for unseen NDRAs. Figure~\ref{fig:class_distribution} displays the newly annotated classes and their distributions. We observe that the class distributions are severely imbalanced, with 84.7\% of the training set focused on normal driving, and the remaining 15.3\% allocated to the eight NDRA classes. For this reason, we use the \textit{mean average precision} (mAP) as the metric for evaluating the models' performance.

\noindent \textbf{Training.} The videos in the DAD dataset have nearly twice the frame rate of commonly used human action datasets, so we remove every other frame to reduce computation. For the temporally downsampled dataset, we construct input clips in the following way. From every 32 frames, 8 of them with equal spacing are randomly selected to introduce temporal scale variation. Then, we leverage the same techniques (e.g., cropping, flipping) to augment the 8-frame clip, which is subsequently resized to the spatial size $112 \times 112$. The R3D-18 backbones \cite{tran2018closer} are pre-trained on Kinetics-400 \cite{carreira2017quo}, and the fusion modules and the MLP are randomly initialized. We train each model using the proposed SuMoCo framework with a queue size of 16,384, a temperature 0.07, and a momentum 0.999 \footnote{The temperature is $\tau$ in \eqref{eqn:10}, and the momentum is $m$ in \eqref{eqn:11}. The choice of these values follow MoCo \cite{he2020momentum}.}. An Adam algorithm \cite{kingma2014adam} with the initial learning rate 1e-3 and a cosine annealing scheduler \cite{loshchilov2016sgdr} are employed to optimize the parameters. Each model is trained with a batch size 32 for 50 epochs.

\begin{table}[t]
\definecolor{shade}{HTML}{ECF4FF}
\definecolor{best}{HTML}{FE0000}
\sethlcolor{shade}

\resizebox{\columnwidth}{!}{%
\begin{tabular}{l|cccccc}
\hline
\textbf{Source}        & \textbf{Decision} & \textbf{Sum} & \textbf{Conv} & \textbf{SE} & \textbf{AFF} & \textbf{MHSA (ours)}                  \\ \hline
\textbf{Top (D)}       & \multicolumn{6}{c}{84.3}                                                                                              \\
\textbf{Top (IR)}      & \multicolumn{6}{c}{83.7}                                                                                              \\
\textbf{Top (D+IR)} &
  84.5 &
  \cellcolor{shade}85.0 &
  \cellcolor{shade}85.4 &
  \cellcolor{shade}85.4 &
  \cellcolor{shade}85.4 &
  \cellcolor{shade}\textbf{85.7} \\ \hline
\textbf{Front (D)}     & \multicolumn{6}{c}{87.7}                                                                                              \\
\textbf{Front (IR)}    & \multicolumn{6}{c}{83.7}                                                                                              \\
\textbf{Front (D+IR)} &
  87.9 &
  87.7 &
  \cellcolor{shade}88.1 &
  \cellcolor{shade}88.2 &
  \cellcolor{shade}88.5 &
  \cellcolor{shade}\textbf{88.7} \\ \hline
\textbf{Top+Front (D)} & 90.7              & 90.1         & 90.4          & 90.5        & 90.6         & \cellcolor{shade}\textbf{90.9} \\
\textbf{Top+Front (IR)} &
  88.4 &
  \cellcolor{shade}89.9 &
  \cellcolor{shade}90.2 &
  \cellcolor{shade}90.2 &
  \cellcolor{shade}90.4 &
  \cellcolor{shade}\textbf{90.6} \\
\textbf{Top+Front (D+IR)} &
  90.9 &
  90.8 &
  \cellcolor{shade}91.2 &
  \cellcolor{shade}91.4 &
  \cellcolor{shade}91.5 &
  \cellcolor{shade}{\color{best} \textbf{91.6}} \\ \hline
\end{tabular}%
}
\caption{The mAP scores for multi-classification of drivers' activities on DAD. The best scores for each view and modality are in \textbf{bold}, and scores with the \hl{blue} background indicate that the corresponding feature-level fusion strategy is better than the decision-level fusion under the metric of mAP.}
\label{tbl:mutiple}
\end{table}

\subsection{Single View/Modality Cases}

In this section, we compare the proposed contrastive learning framework SuMoCo with the CL method in \cite{kopuklu2021driver} for single-view single-modal NDRA detection (binary classification between normal driving and NDRA). We report the AUC-ROC scores in Table \ref{tbl:sumoco}, where the second row shows the results of SuMoCo trained with the same binary labels as Kopuklu \textit{et al}. \cite{kopuklu2021driver}. We observe that SuMoCo outperforms Kopuklu \textit{et al}. by non-trivial margins ($>1.5\%$) on the depth modality and by smaller margins ($\le 0.5\%$) on IR. These results demonstrate the effectiveness of SuMoCo, which has fewer contrastive pairs in each batch to save GPU memory.

Our new annotations enable SuMoCo to be trained for multi-classification, and we report its detection performance by comparing the score for normal driving and the sum of scores for NDRAs (results shown in the last row of Table \ref{tbl:sumoco}). Our annotations can further improve the performance of SuMoCo (up to $2.2\%$), as the more detailed label information can regularize the contrastive embedding space. Starting from here, we train our models with the new labels for multi-class classification and also report their performances in NDRA detection.

\subsection{Multiview Multimodal Fusion}

Table \ref{tbl:binary} compares our feature-level fusion models with the decision-level approach proposed in \cite{kopuklu2021driver} for NDRA detection. Our multi-source DMS with the MHSA fusion module consistently outperforms all other methods, demonstrating the superior performance of this self-attention mechanism in multi-view/modal fusion. The highest ROC and mAP scores ($97.0\%$ and $97.8\%$, respectively) are achieved when all four data sources are combined. Additionally, MHSA is the most stable method for this task, as its performance consistently improves when an extra view/modality is introduced.

We remove the unseen NDRAs in the test set to evaluate our models' performance on multi-class classification, since our work does not focus on open-set recognition. Table \ref{tbl:mutiple} shows mAP scores. We observe that the MHSA-based model again outperforms all other fusion methods, with the highest scores ($91.6\%$) achieved when combining two views and two modalities. Table \ref{tbl:mutiple} also indicates that the mAP scores of our proposed fusion models can always be improved as more sources are included, demonstrating their effectiveness in multi-view multi-modal action recognition.

As for the importance of each modality, we find all models consistently perform better on the depth modality than on the IR modality. Furthermore, having two views is more beneficial than having two modalities, as the models have larger performance improvements. These findings suggest that the top depth and front depth data sources are the most useful in practice, as DMS built on them can achieve good performance while enjoying relatively low computation costs.

\begin{table}[tb]
\centering
\begin{tabular}{l|cc}
\toprule
\textbf{Fusion} & \textbf{1-Step} & \textbf{2-Step} \\ \midrule
Sum             & 90.8            & \textbf{91.4}   \\
Conv            & \textbf{91.2}   & 91.1            \\
SE              & 91.4            & \textbf{91.6}   \\
AFF             & 91.5            & \textbf{91.6}   \\
\bf MHSA  (Ours)           & \textbf{91.6}   & 90.9            \\ \bottomrule
\end{tabular}
\caption{Comparisons (in mAP) between 1-step fusion (all at once) and 2-step fusion of the four input sources in DAD on multi-classification. Results show that fusing all views/modalities at the same step is beneficial for MHSA.}
\label{tbl:1-vs-2}
\end{table}

\begin{figure}[tb]
    \centering
    \begin{subfigure}{0.495\linewidth}
        \centering
        \includegraphics[width=\textwidth]{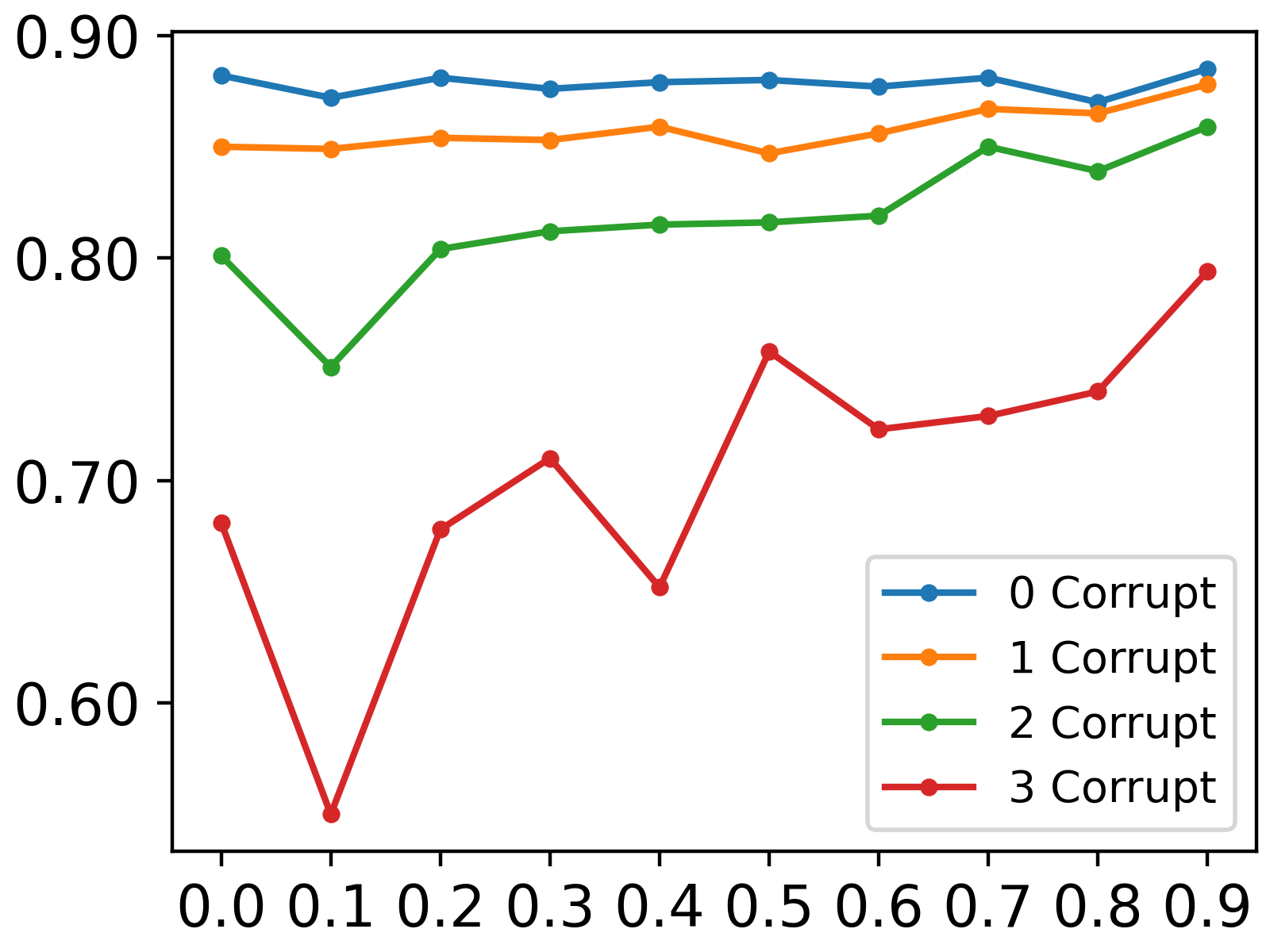}
        \caption{Mult. Acc.}
        \label{fig:robust_cls_acc}
    \end{subfigure}
    \hfill
    \begin{subfigure}{0.495\linewidth}
        \centering
        \includegraphics[width=\textwidth]{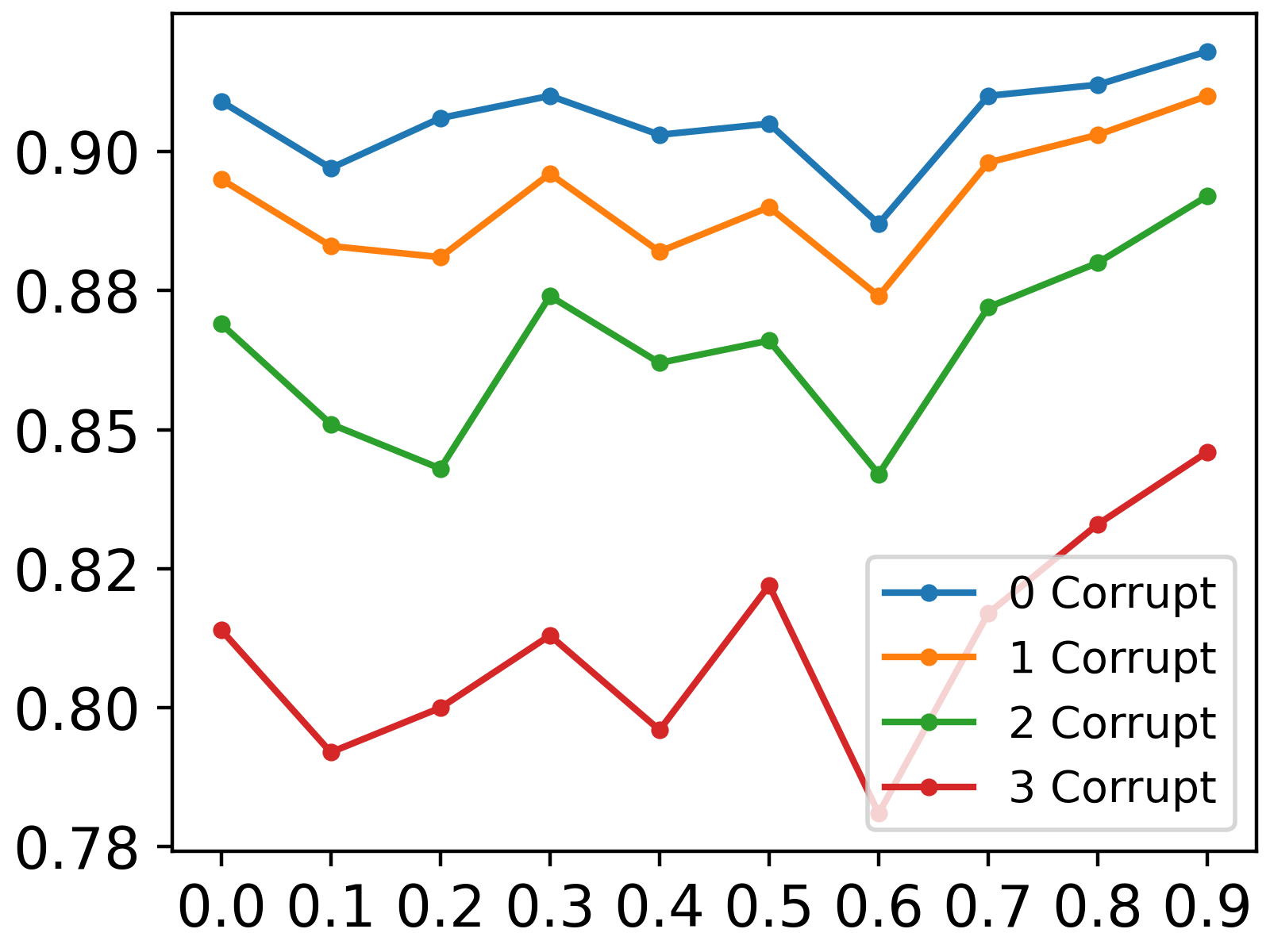}
        \caption{Bin. Acc.}
        \label{fig:robust_dtc_acc}
    \end{subfigure}
    \begin{subfigure}{0.495\linewidth}
        \centering
        \includegraphics[width=\textwidth]{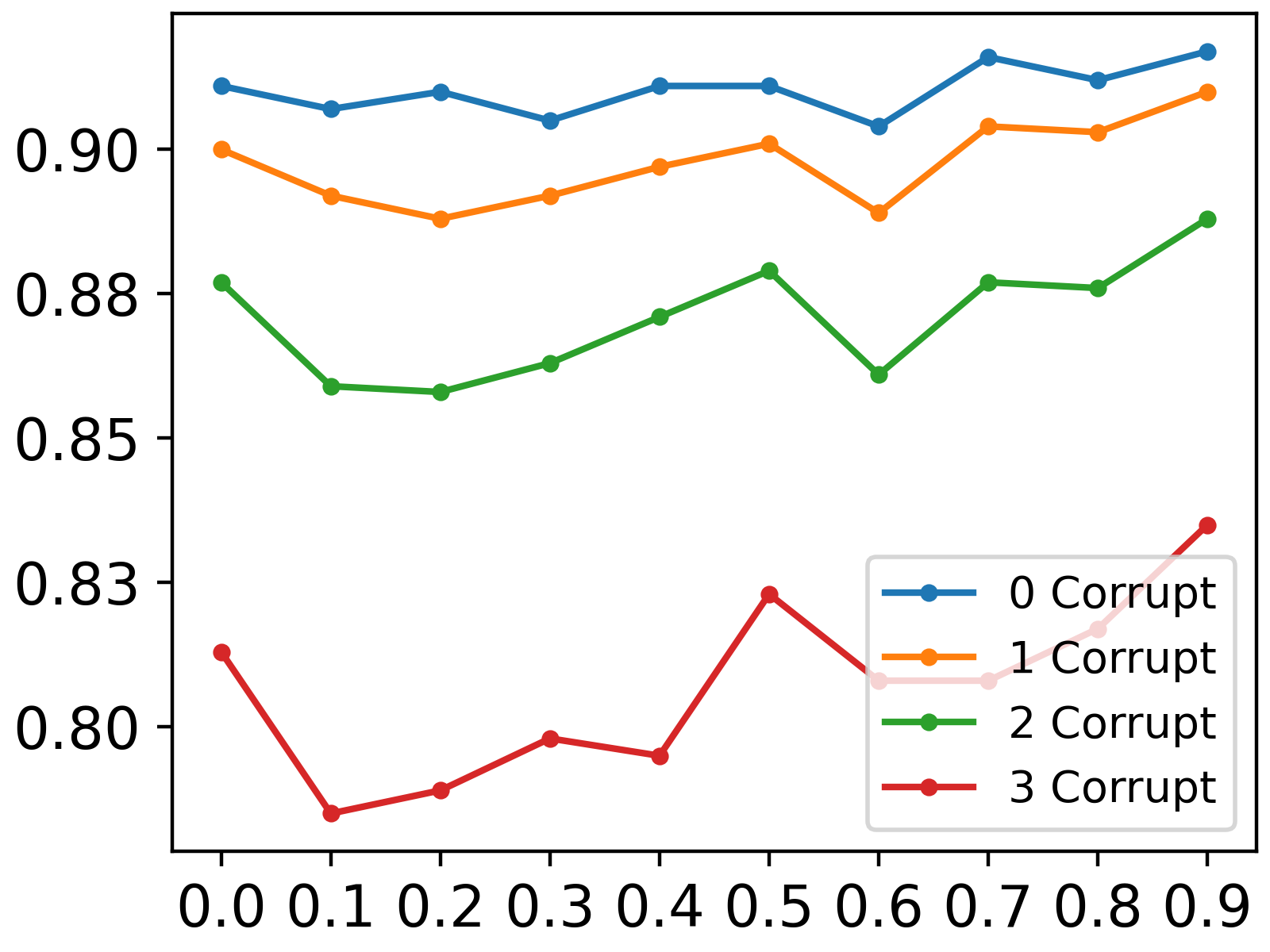}
        \caption{Mult. mAP}
        \label{fig:robust_cls_pr}
    \end{subfigure}
    \hfill
    \begin{subfigure}{0.495\linewidth}
        \centering
        \includegraphics[width=\textwidth]{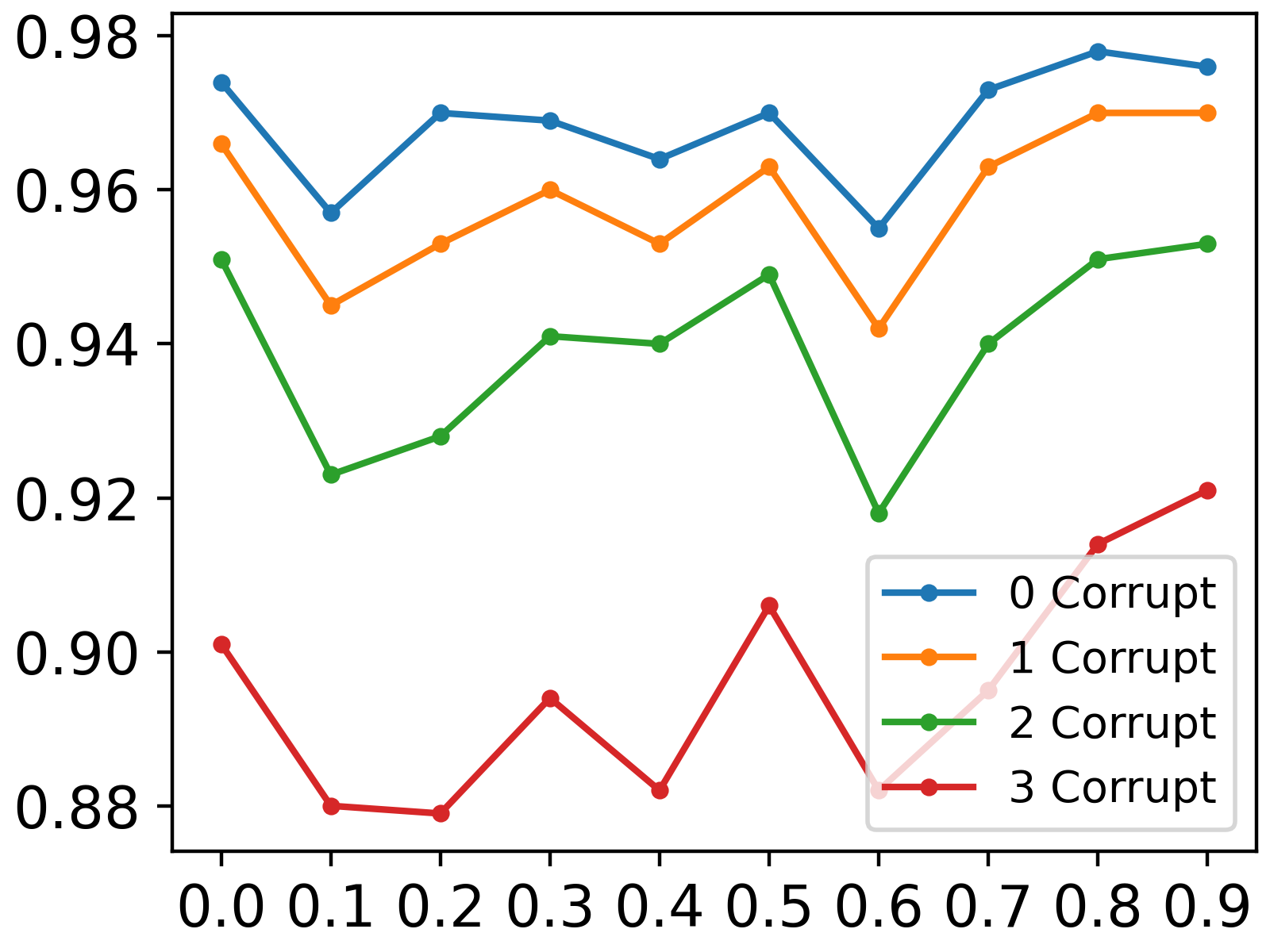}
        \caption{Bin. mAP}
        \label{fig:robust_dtc_pr}
    \end{subfigure}
    \caption{Masked training improves MHSA's robustness against corrupt views/modalities. MHSA is trained with all four data sources in DAD and a varying mask ratios ranging from $0.0$ (i.e., no masking) to $0.9$ (i.e., $90\%$ of the patches are masked). In testing, it is evaluated with zero to three data sources collapsed. Thus, small performance degradation indicates stronger robustness against corrupt data sources. The $x$-axis in the resulting plots indicates the mask ratio, and the $y$-axis displays the corresponding average score.}
    \label{fig:robust}
\end{figure}

\subsection{One-Step or Two-Step?}

In this section, we investigate whether views and modalities should be combined altogether in a single step, or in separate steps. The previous results are based on the \textit{one-step} fusion, where feature maps from all four sources ( $\boldsymbol{f}_\text{F, D}$, $\boldsymbol{f}_\text{F, IR}$, $\boldsymbol{f}_\text{T, D}$, and $\boldsymbol{f}_\text{T, IR}$) are simultaneously fused. In contrast, the \textit{two-step} strategy fuses features sequentially. We fuse the features from the same views first to ensure spatial consistency: $\boldsymbol{f}_\text{F} = \text{fusion}_\text{F} ( \boldsymbol{f}_\text{F, D}, \, \boldsymbol{f}_\text{F, IR})$ and $\boldsymbol{f}_\text{T} = \text{fusion}_\text{T} ( \boldsymbol{f}_\text{T, D}, \, \boldsymbol{f}_\text{T, IR})$. Then $\boldsymbol{f}_\text{F}$ and $\boldsymbol{f}_\text{T}$ are further fused.

Table \ref{tbl:1-vs-2} compares the results of the two fusion types. We observe that for Sum, SE, and AFF, the two-step fusion outperforms the one-step fusion, while for Conv, the results are similar. The two-step fusion method is easier to learn because the input features have more similar semantics than those of one-step fusion. However, for MHSA, fusing features in two steps leads to a degradation in performance due to overfitting – the number of transformer encoder blocks is increased from two to four.

 \subsection{Patch Masking for Robustness}

In this section, we validate that training the MHSA fusion method with masking can improve its robustness against modality/view collapses. MHSA is trained on the four data sources with the mask ratio varying from $0.0$ (no masking) to $0.9$ (90\% of patches are randomly removed). In the test time, we collapse one, two, and three data sources and feed the remaining to the model. For each collapse scenario, we calculate the scores and average them over the number of possible collapses. For instance, there are $C^2_4 = 6$ choices for two corrupt sources, so we calculate the score for each case and then average them. Figure \ref{fig:robust} illustrates these results. With the mask ratio increasing, the performance of MHSA in all collapse scenarios also show an upward trend. The MHSA with the mask ratio $0.9$ is the most robust as the differences of its test performances are the slightest. These observations show that random masking can improve the robustness against missing sources. 

Moreover, we find that when there is no collapse, the scores still exhibits an increasing trend as the mask ratio increases, showing that random masking, as a strong regularization technique, may improve the model's generalization capabilities. This finding coincides with the result of two-step fusion (i.e., MHSA may overfit the training set). Overall, our results suggest that training MHSA with random masking can improve its generalization capabilities and robustness against modality/view collapses.

\begin{figure}[tb]
    \centering
    \includegraphics[width=\linewidth]{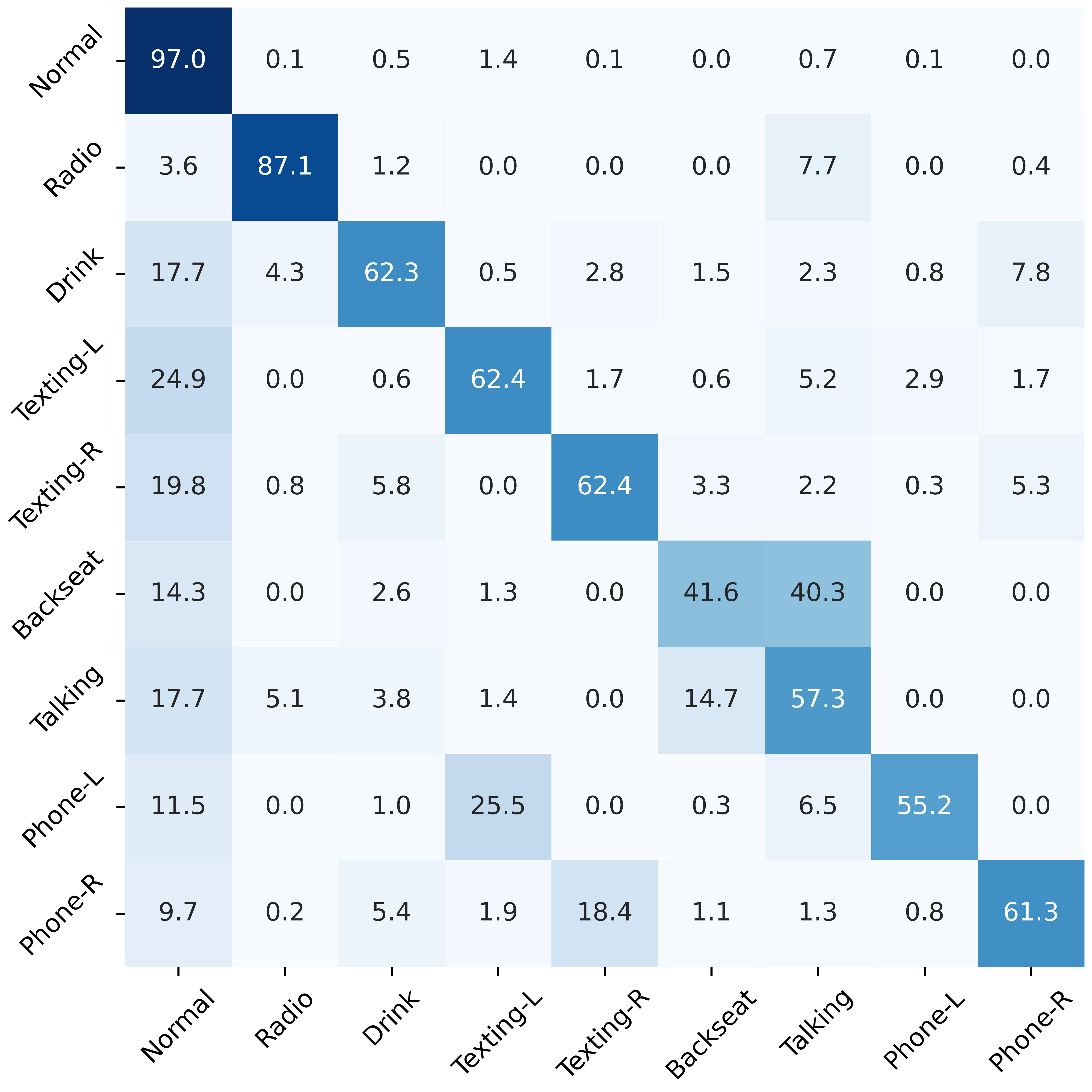}
    \caption{The confusion matrix for acitivity classification on DAD using the MHSA model. Each element is normalized by its row sum, so the diagonal entries represent the recall values (in percent) for class.}
    \label{fig:cm}
\end{figure}

\subsection{Confusion among Classes}

Figure \ref{fig:cm} depicts the confusion matrix of our MHSA model. We find that some NDRAs are misclassified as normal driving. This issue can be partially attributed to class imbalance, as each NDRA class only constitutes 1.9\% of the training set while normal driving comprises 84.7\%. This heavy imbalanced distribution makes our model overfit to the normal driving data. For ``tuning the radio'', our model can recognize most of this action (87.1\%), while for other NDRA, the recognition rates are not very comparable. Upon further scrutiny of the test set, we find that this problem is caused by untrimmed video clips of other classes. For example, the action of talking on the phone with the right hand in \texttt{val03} \texttt{rec1} is annotated to start from frame 8467, but for us this action can only be recognizable from frame 8760 (about 6.2s later). The last row of Fig. \ref{fig:vis}, which corresponds to frame 8516, illustrates this case. These frames, which are originally labeled as an NDRA by \cite{kopuklu2021driver} but are actually normal driving, leads to the fake errors in Fig. \ref{fig:cm}. Interestingly, the model seems very confused between reaching behind and talking to the passenger, probably because both actions need drivers to turn their heads back. Also, the DMS sometimes confuses between texting and talking on the phone, which is not beyond expectation as drivers may talk on the phone in the hands-free mode while the DAD dataset has no audio available.

{
\definecolor{myred}{HTML}{FF6666}
\definecolor{mygreen}{HTML}{00CC66}
\begin{figure*}[t]
    \centering
    \includegraphics[width=\textwidth]{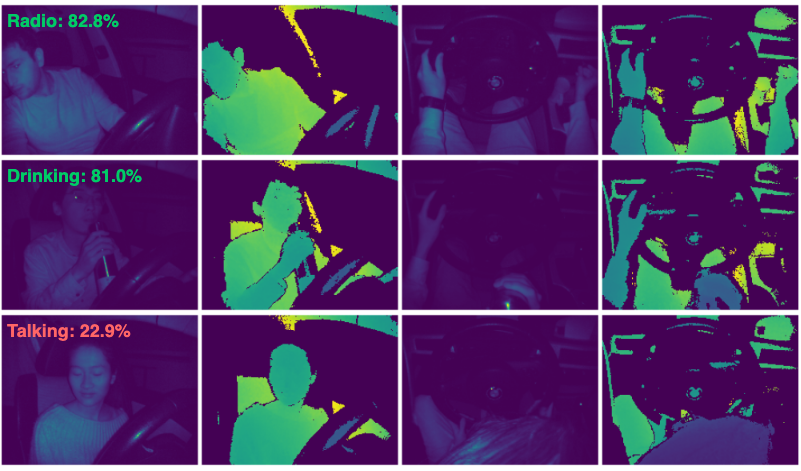}
    \caption{Visualizaion of the middle frames of four test samples from DAD. From left to right, the data sources are front IR, front depth, top IR and top depth, respectively. The text at the upper left corner indicates the predicted class of the driver's activity and the corresponding probability. Those in \textcolor{mygreen}{green} indicate that the predictions are correct. Our proposed method MHSA makes a fake error for the last case (in \textcolor{myred}{red}). It predicts the NDRA to be talking to the passenger, but the actual label is talking on the phone.}
    \label{fig:vis}
\end{figure*}
}

\section{Discussions and Conclusions}

In this paper, we proposed (i) a novel multiview multimodal driver monitoring system (DMS) with an effective fusion strategy based on multi-head self-attention (MHSA) and (ii) a GPU-friendly supervised contrastive learning framework, SuMoCo. In addition, we also labeled the DAD dataset (which initially had labels for binary classification) to enable multi-class classification. Extensive experiments on the DAD dataset verified the effectiveness of our proposed methods. We demonstrated that the MHSA-based fusion strategy achieves the best results compared to other competitive fusion methods on both NDRA detection and action classification. We also showed that training MHSA with random patch masking can enhance its robustness against missing input channels (modality/view).

\noindent\textbf{Limitations.} One drawback of our fusion model is that it overfits the training set. For future work, we will study the feasibility of using a single branch to handle all data sources to restrict the model's expressivity, thereby reducing overfitting. We will also collect more data for non-driving-related activities (NDRAs) to resolve the data imbalance issue. 

\noindent\textbf{Negative Impacts.} Our work is evaluated on a public dataset that is not balanced regarding ethnicity, religion, and other demographic factors. We will consider these factors when we collect data in the future. Also, data about drivers' faces are required for our DMS, since it needs to detect activities like talking. This system may thus be misused to potentially infer a person's identity-related information.

{\small
\bibliographystyle{ieee_fullname}
\bibliography{egbib}
}

\end{document}